\documentclass{article} 
\usepackage{iclr2026_delta,times}

\usepackage{amsmath,amsfonts,bm}

\def\eqref#1{equation~\ref{#1}}

\def\1{\bm{1}}

\DeclareMathAlphabet{\mathsfit}{\encodingdefault}{\sfdefault}{m}{sl}
\SetMathAlphabet{\mathsfit}{bold}{\encodingdefault}{\sfdefault}{bx}{n}

\usepackage{hyperref}
\usepackage{url}

\usepackage{algorithm}
\usepackage{algorithmic}
 \usepackage{booktabs}

\usepackage{amsmath}
\usepackage{amssymb}
\usepackage{mathtools}
\usepackage{amsthm}
\usepackage{subcaption}

\theoremstyle{definition}
\newtheorem{definition}{Definition}
\iclrfinalcopy

\title{Generative Hints}

\author{
Dimnaku, Andy\thanks{Work done while at California Institute of Technology.} \\
Department of Electrical Engineering \\
Stanford University \\
Palo Alto, CA 94305, USA \\
\texttt{adimnaku@stanford.edu} \\
\And
Kavranoglu, A. Yusuf \\
Department of Electrical Engineering \\
California Institute of Technology \\
Pasadena, CA 91125, USA \\
\texttt{ykavranoglu@gmail.com} \\
\And
Abu-Mostafa, Yaser \\
Department of Electrical Engineering \\
California Institute of Technology \\
Pasadena, CA 91125, USA \\
\texttt{yaser@caltech.edu} \\
}

\begin{document}

\maketitle

\begin{abstract}
Data augmentation is widely used in vision to introduce variation and mitigate overfitting, by enabling models to learn invariant properties. However, augmentation only indirectly captures these properties and does not explicitly constrain the learned function to satisfy them beyond the empirical training set. We propose \textit{generative hints}, a training methodology that directly enforces known functional invariances over the input distribution. Our approach leverages a generative model trained on the training data to approximate the input distribution and to produce \textit{unlabeled} synthetic images, which we refer to as \textit{virtual examples}. On these virtual examples, we impose hint objectives that explicitly constrain the model’s predictions to satisfy known invariance properties, such as spatial invariance. Although the original training dataset is fully labeled, generative hints train the model in a semi-supervised manner by combining the standard classification objective on real data with an auxiliary hint objectives applied to unlabeled virtual examples. Across multiple datasets, architectures, invariance types, and loss functions, generative hints consistently outperform standard data augmentation, achieving accuracy improvements of up to 2.10\% on fine-grained visual classification benchmarks and an average gain of 1.29\% on the CheXpert medical imaging dataset.
\end{abstract}

\section{Introduction}

Data augmentation has become ubiquitous in computer vision, with transformations such as rotations, crops, and color jittering serving as essential components of modern training pipelines \citep{shorten2019survey}. By exposing models to transformed versions of training examples, augmentation helps models implicitly learn invariance to these transformations. For instance, by training on both an original image and its horizontally flipped version with the same label, a classifier learns that predictions should remain unchanged under horizontal flips. This implicit learning of invariances has proven crucial for achieving state-of-the-art performance across diverse vision tasks \citep{perez2017effectiveness}.
  
Despite its widespread success, data augmentation has two fundamental limitations. First, it only implicitly encourages invariance through exposure to transformed examples, without explicitly enforcing that the model's predictions satisfy invariance constraints. The model may learn to be invariant on augmented training examples while still violating invariance properties on unseen data from the input distribution. Second, augmentation is inherently limited to the finite training set, whereas the true input distribution is continuous and extends beyond observed examples. These limitations motivate the need for approaches that can explicitly enforce invariances over the learned input distribution.
  
We build on the concept of \textit{hints} from \citep{AbuMostafa1990Hints}, which encode known properties of the target function as auxiliary training objectives. In the original formulation for tabular data, hints were enforced on random samples from simple noise distributions (e.g., uniform or Gaussian noise), which sufficiently covered the low-dimensional input space. However, in high-dimensional vision tasks, random noise lies far from the natural image manifold, rendering this strategy ineffective. To address this, we leverage modern generative models to approximate the input distribution and sample realistic synthetic images as \textit{virtual examples} (following the terminology of \citep{AbuMostafa1995Hints}) on which to enforce hints.

Our approach introduces an auxiliary training objective on synthetic data: while the classification loss operates on labeled training examples, the hint loss operates on unlabeled virtual examples sampled from a generative model trained on the training data. For instance, to enforce spatial invariance, we sample a virtual example $x_v$ from the generative model, apply a spatial transformation $h(x_v)$ (such as rotation or translation), and minimize the divergence between the model's predictions on $x_v$ and $h(x_v)$. This explicitly enforces that the model's learned function satisfies $f(x_v) = f(h(x_v))$ over the distribution approximated by the generative model, rather than just on the finite training set used by standard augmentation.

Crucially, generative hints compound with standard data augmentation rather than replacing it. Both our baseline and hint-based training use identical augmentation strategies on labeled training data; the only difference is that hint-based training additionally enforces invariance constraints on unlabeled virtual examples. This controlled design allows us to isolate the contribution of explicit invariance enforcement. We demonstrate that this combination consistently outperforms augmentation-only baselines across diverse settings. Specifically, Our key contributions are:

\begin{enumerate}

    \item We demonstrate that generative hints consistently improve over data augmentation baselines across multiple architectures (ViT-B, Swin-B, ResNet50), datasets (Stanford Cars, FGVC Aircraft, CUB-200-2011, Oxford Flowers, CheXpert), and loss functions (cross-entropy and MSE). Hints achieve accuracy gains of up to 2.10\% on fine-grained visual classification and an average improvement of 1.29\% on CheXpert medical imaging.

     \item To our knowledge, we are the first to reformulate a fully supervised visual classification task with labeled training data as a semi-supervised training problem by treating data synthesized from a generative model as \textit{unlabeled data}.

     \item We introduce a training methodology that explicitly enforces known invariance properties of the target function over the learned input distribution, rather than relying solely on implicit learning through data augmentation on the finite training set.
     
\end{enumerate}

\section{Related Work}
\subsection{Generative Models for Vision}

\textbf{Generative Models} Recent advances in generative modeling have enabled the synthesis of high-fidelity images from noise, primarily through diffusion models \citep{ho2020denoising, song2020score, rombach2022high} and GANs \citep{goodfellow2014generative, karras2019stylegan, karras2020stylegan, karras2021stylegan}. These models have been applied both as tools for data generation and as components of downstream training pipelines. In classification, discriminators have been adapted for semi-supervised learning \citep{kingma2014semi, radford2015unsupervised}, while synthetic data has been used to expand training sets in medical and natural image domains \citep{antoniou2017dagan, fridadar2018liver, zhao2019transform, azizi2023synthetic, yuan2024realfake}. More recently, diffusion-based pipelines \citep{bordes2023diffusion, huang2023ttida, zhang2024aga} highlight the ability of generative models to provide controllable, task-aware augmentation. However, these approaches primarily focus on increasing data diversity rather than directly enforcing functional properties.

\textbf{Data Augmentation and Invariances} Conventional data augmentation is widely used to induce invariances (e.g., spatial or color invariance) by perturbing training examples. While effective for regularization, this strategy only encourages models to learn invariance indirectly, relying on the hope that augmented samples approximate invariance-preserving transformations \citep{perez2017effectiveness, shorten2019survey}.

\textbf{Generative Data Augmentation} Generative data augmentation (GDA) builds on generative models to synthesize additional labeled data, with demonstrated benefits in low-data regimes, domain-specific applications, and joint generation–classification frameworks \citep{mahapatra2022unified}. Yet, existing GDA methods treat generated examples primarily as extra training data, without using them to explicitly encode known invariances or functional constraints.

Unlike standard augmentation or GDA, generative hints use synthetic examples as unlabeled carriers of functional properties. That is, generative hints focus is on learning properties of the target function through our semi-supervised training so it can be additively done with existing GDA works. 

\subsection{Hints}

\textit{Hints} were first introduced in \citep{AbuMostafa1990Hints, AbuMostafa1995Hints} to teach machine learning models functional properties of the target function and data. These properties, referred to as hints, are incorporated as auxiliary objectives optimized alongside the main task. For example, in credit default prediction using tabular data, the target is to predict whether a default will occur given input features. Domain knowledge provides that, if all other features remain fixed while income increases, the probability of default should decrease. This property can be formalized as a \textit{monotonicity hint} and enforced through an auxiliary loss. Similarly, in the foreign exchange (FX) markets, a \textit{symmetry hint} has been used to regularize models against noisy financial data, leading to significantly improved annualized returns. Generative Hints is explicitly different from previous iterations of hints in its formulation of using a generative model to represent the input space to learn the functional properties.

\section{What are Hints?}

\subsection{Problem Statement}

We begin by defining the standard supervised learning setup. Let$f: X \rightarrow Y$ denote the true underlying function, where $X$ is the input space and $Y$ is the output space. Let $D_{train}$ and \textit{$D_{test}$} denote the training and test sets, respectively. In the case of image classification, $X$ corresponds to the space of images and $Y$ to the space of class probability distributions.

Intuitively, a \textit{hint} encodes domain knowledge about a target function $f$ by specifying how the output should behave under certain input transformations. For instance, in image classification, we know that flipping an image horizontally should not change its class label distribution. Rather than learning such properties implicitly form labeled data, hints allows us to enforce them explicitly during training. We now formalize this notion, beginning with the general concept of hints and the specializing with invariance hints.

\begin{definition}[Invariance Hint]
A \textit{hint} is a constraint on the target function $f$ expressed through input transformations. Formally, let $h: X \rightarrow X$ be a transformation function and $R: Y \times Y \rightarrow \{0,1\}$ be a binary relation on the output space. A hint specifies that for any input $x \in X$ the relationship $R(f(x), f(h(x)))=1$ must hold. An \textit{invariance hint} is a hint where the relationship $R$ enforces output equality, specifying that for any $x \in X$, 
\[f(h(x))=f(x).\]
\end{definition}

While data augmentation and our hint-based approach both leverage invariance, they differ fundamentally in their mechanism. Data augmentation applies transformations to \textit{labeled} training examples $(x,y) \in D_{train}$, creating additional training pairs $(h(x),y)$ and rely on the supervised loss to implicitly teach invariance. In contrast, our method applies hints to \textit{unlabeled} virtual examples (defined in Section 3.2) and uses an explicit auxiliary loss to enforce the invariance property directly. In practice, we apply both techniques simultaneously, with data augmentation operating on the labeled training data and hints providing additional invariance signal through virtual examples.

\subsection{Enforcing Hints through Virtual Examples}

We enforce invariance hints on \textit{virtual examples}, which are unlabeled synthetic inputs from a generative model trained on the input distribution. Virtual examples were originally introduce in \citep{AbuMostafa1995Hints} for tabular data, where they were sampled from Gaussian noise. We adapt this concept to high-dimensional vision tasks by sampling from a learned generative model, ensuring the virtual examples are realistic and representative of the input image distribution.

\begin{definition}[Invariance Hint on Virtual Examples]
Given a generative model $G_{\theta}$ and transformation $h: X \rightarrow X$, an \textit{invariance hint on virtual examples} enforces that for any $z\sim p(z)$, 
\[ f(h(G_\theta(z))) = f(G_\theta(z)). \]
\end{definition}

We instantiate our framework with three invariance hints commonly leveraged in data augmentation, corresponding to properties that image classification naturally respect.

\begin{definition}[Spatial Invariance Hint]
Let $h_{\text{sp}}: X \rightarrow X$ denote a spatial transformation that applies a random horizontal flip with probability $p_{flip}$, translation within $\pm t$ pixels, and rotation within $\pm r$ degrees. A \textit{spatial invariance hint} enforces that for any virtual example $x_v$ sampled from $G_{\theta}$, 
\[
f(h_{\text{sp}}(x_v)) = f(x_v).
\]
\end{definition}

\begin{definition}[Photometric Invariance Hint]
Let $h_{\text{ph}}: X \rightarrow X$ denote a photometric transformation that adjusts brightness, contrast, and saturation by a multiplicative factor sampled uniformly from $[1-\alpha, 1+\alpha]$, where $\alpha > 0$ controls the strength of the transformation.  
A \textit{photometric invariance hint} enforces that for any virtual example $x_v$ sampled from $G_{\theta}$, 
\[
f(h_{\text{ph}}(x_v)) = f(x_v).
\]
\end{definition}

\begin{definition}[Cropping Invariance Hint]
Let $h_{\text{cr}}: X \rightarrow X$ denote a random crop operator that extracts a subregion of size $b \times b$ from an $a \times a$ image, where $b < a$. A \textit{cropping invariance hint} enforces that for any virtual example $x_v$ sampled from $G_{\theta}$, \[
f(h_{\text{cr}}(x_v)) = f(x_v).
\]
\end{definition}

\subsection{Training Generative Models Efficiently}

We use StyleGAN3 from \cite{karras2021stylegan} as our generative model to produce virtual examples. This model generates unlabeled images from the input distribution without class conditioning. While other generative models (ex. diffusion models) could be used, StyleGAN3 provided a favorable balance between sampling efficiency and image quality.

Training generative models in limited data settings requires careful data-efficient strategies to prevent overfitting. We leverage adaptive discriminator augmentation (ADA) from \cite{Karras2020ada}, which adjusts the strength of data augmentations dynamically based on overfitting signals, improving image quality in low-data regimes.We extend ADA with a curriculum learning approach where the full details can be seen in the Appendix. 

\subsection{Hint Loss Function}

To enforce invariance, we measure prediction similarity using symmetric KL divergence with temperature scaling. For virtual example $x_v$ with predictions $p = \hat{f}(x_v)$ and $q = \hat{f}(h(x_v))$, the temperature-scaled distributions are $\tilde{p} = \mathrm{softmax}(p/T)$ and $\tilde{q} = \mathrm{softmax}(q/T)$. The hint loss is:
\[ \mathcal{L}_{\text{hint-ce}}(\tilde{p}, \tilde{q}) = \frac{1}{2} \Big( \mathrm{KL}(\tilde{p} \,\|\, \tilde{q}) + \mathrm{KL}(\tilde{q} \,\|\, \tilde{p}) \Big). \]

For MSE-based training, we use: $\mathcal{L}_{\text{hint-mse}}(y_v, y'_v) = \frac{1}{d} \sum_{i=1}^{d} (y_{v,i} - y'_{v,i})^2$.

\subsection{Training Algorithm}

Our approach optimizes both classification loss on labeled data from $D_{train}$ and hint loss on unlabeled virtual examples from $G_{\theta}$, alternating between objectives each batch (Algorithm 1). The virtual examples are generated on-the-fly, ensuring diversity.

\begin{algorithm}[tb]
  \caption{Generative Hints Training Algorithm}
  \label{alg:gen-hints}
  \begin{algorithmic}
    \STATE {\bfseries Input:} Training set $\mathcal{D}_{\text{train}}=\{(x_i,y_i)\}_{i=1}^N$
    \STATE Classifier $\hat f$, hint transformation $h: X \rightarrow X$, generator $G_{\theta}$
    \STATE Losses $\mathcal{L}_{\text{class}}$, $\mathcal{L}_{\text{hint}}$, coefficient $\alpha$
    \STATE Number of epochs $E$
    \FOR{epoch $e=1$ {\bfseries to} $E$}
      \FOR{mini-batch $B \subset \mathcal{D}_{\text{train}}$}
        \STATE Update $\hat f$ on $B$ using $\mathcal{L}_{\text{class}}$ applying transformation $h$
        \STATE Sample $z \sim p(z)$ and set $x_v \sim G_{\theta}(z)$
        \STATE Compute $x_v' y_v \gets \hat f(x_v)$ and $y_v' \gets \hat f(x_v')$
        \STATE Update $\hat f$ using $\alpha \mathcal{L}_{\text{hint}}(y_v,y_v')$
      \ENDFOR
    \ENDFOR
  \end{algorithmic}
\end{algorithm}

We introduce a coefficient $\alpha$ to scale the hint loss, controlling its relative weight compared to the classification objective. This weighting is necessary because the gradients and learning dynamics of the two objectives can differ significantly. In our experiments, we used a fixed $\alpha$ which provides stable and consistent improvements across datasets and architectures.

\section{Experiments and Results}

We evaluated our method on four popular fine-grained visual classification datasets: Stanford Cars \citep{krause2013collecting}, CUB-200-2011 \citep{WahCUB_200_2011}, FGVC Aircraft \citep{maji13fine-grained}, and Oxford Flowers \citep{nilsback2008flowers}. Experiments were conducted using two model architectures: ViT-B \citep{vit, vaswani2017attention} and Swin-B \citep{swint}, chosen for their strong performance on fine-grained classification and to demonstrate the generality of our approach across architectures. We evaluated three hint types: spatial invariance, photometric invariance, and croppign invariance.

We further evaluated generative hints in a medical imaging setting using the CheXpert dataset \citep{irvin2019chexpert} with a ResNet50 \citep{resnet}, employing mean squared error as the training objective. Finally, we performed ablation studies to examine the impact of generative model quality on classification performance. All experiments were conducted on a single NVIDIA H100 GPU, training both the generative and classification models.

\subsection{Generative Model Training}

\begin{figure}[t]
    \centering
    \includegraphics[width=1.0 \linewidth]{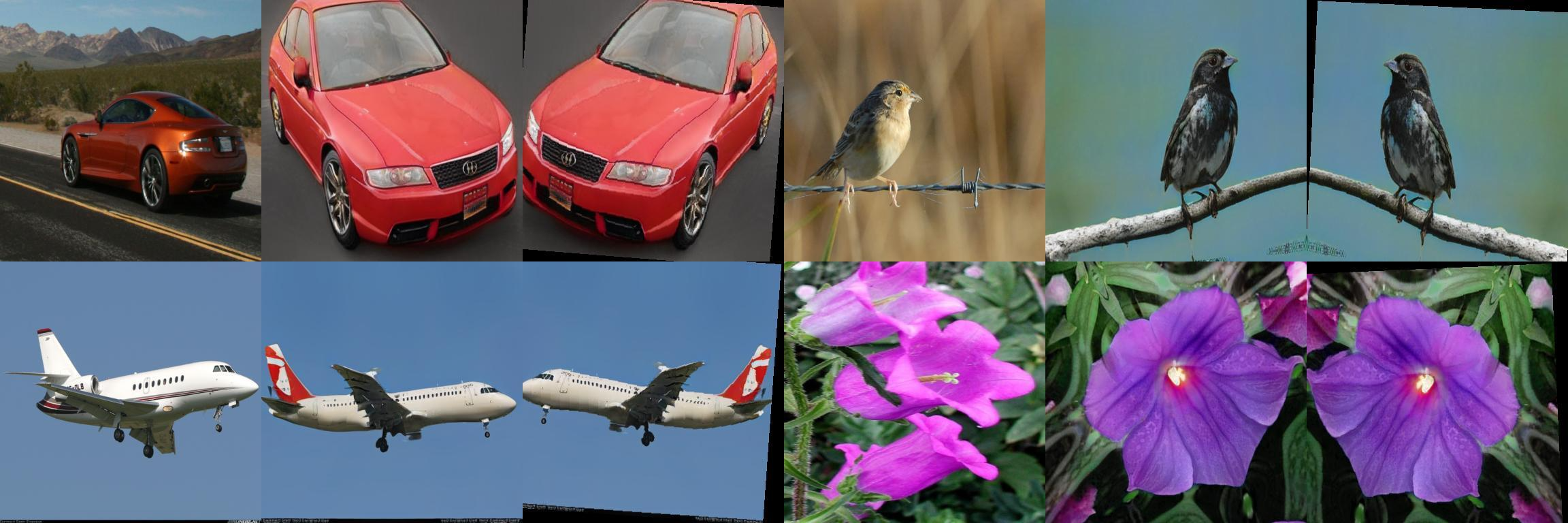}
    \caption{
        Virtual examples for each dataset: Stanford Cars (top left), CUB-200-2011 (top right), FGVC Aircraft (bottom left), and Oxford Flowers (bottom right). Each panel shows (left to right) an original training image from $D_{train}$, a virtual example $x_v$ sampled from the generative model $G_{\theta}$, and the transformed image $h_s(x_v)$ after applying the spatial invariance hint.}
    \label{fig:combined_datasets}
\end{figure}

We trained StyleGAN3 at $512 \times 512$ resolution on each dataset using ADA (see Appendix for hyperparameters). We trained for 5000 kimgs and selected checkpoints with best FID: 5.29 (Stanford Cars), 4.73 (FGVC Aircraft), 7.04 (CUB-200-2011), 12.62 (Oxford Flowers). Figure~\ref{fig:combined_datasets} shows virtual examples from each dataset and Table~\ref{tab:datasets} summarizes the statistics of each dataset and the quality of the trained generative models, measured using Fréchet Inception Distance (FID) \citep{heusel2017gans}.

\begin{table}[h]
    \centering
    \caption{Dataset statistics for the four fine-grained visual classification benchmarks. FID is measured for StyleGAN3 trained on each training set, used for virtual example generation. }
    \begin{tabular}{lrrrr}
        \toprule
        Dataset & Classes & Training Size & FID \\
        \midrule
        Stanford Cars    & 196 & 8,144 & 5.29 \\
        FGVC Aircraft    & 100 & 6,800 & 4.73 \\
        CUB-200-2011     & 200 & 5,994 & 7.04 \\
        Oxford Flowers   & 102 & 2,040  & 12.62 \\
        \bottomrule
    \end{tabular}
    \label{tab:datasets}
\end{table}

\subsection{Fine-Grain Vision Classification Training Results}

\textbf{Experimental Design} To isolate the contribution of generative hints, we carefully control for the effect of data augmentation. Our baseline applies the same transformations used in our hints (spatial, photometric, and cropping) as standard data augmentation during training on labeled data. The generative hints condition uses these same augmentations on the training data, while \emph{additionally} enforcing invariance constraints on unlabeled virtual examples via the hint loss $\mathcal{L}_{\text{hint}}$. This ensures that any performance improvement is attributable to the explicit invariance enforcement on virtual examples rather than the transformations themselves. Both conditions are trained with identical hyperparameters, differing only in the addition of the hint loss on virtual examples.

\textbf{Model and Training Setup} We evaluated ViT-B (patch size 16) and Swin-B (patch size 4), both pretrained on ImageNet-1k. All experiments used $384 \times 384$ resolution, batch size 32, AdamW optimizer (lr=0.0001), and trained for 200 epochs with cosine annealing. Hyperparameters were optimized for baselines using data augmentation and kept fixed for hint training.

\textbf{Augmentation and Hints} Spatial: horizontal flip (50\% baseline, 100\% hints), translation ($\pm 5\%$), rotation ($\pm 5°$). Photometric: brightness/contrast/saturation factors in $[0.8, 1.2]$. Cropping: resize to $448^2$, crop to $384^2$. We used $\alpha=1.0, T=0.8$ (ViT-B) and $\alpha=50.0, T=0.8$ (Swin-B), tuned on Stanford Cars using the photometric hint and fixed for other datasets and hints. Each experiment was carried out with 5 seeds.

For generative hints training, we sampled one virtual example per real training example in each batch from the StyleGAN3 generative model (Section 5.1) and enforced the corresponding hint property. We performed a sweep over the temperature $T \in \{0.5, 0.8, 1.0, 1.2, 1.5, 2.0\}$ and hint loss coefficient $\alpha \in \{1.0, 5.0, 10.0, 25.0, 50.0, 75.0, 100.0\}$, tuning only on the Stanford Cars dataset for the photometric hint. For the final experiments, we used $\alpha=1.0$ and $T=0.8$ for ViT-B, and $\alpha=50.0$ and $T=0.8$ for Swin-B; these values were then kept fixed for all other datasets. Tables~\ref{tab:alpha_comparison} and ~\ref{tab:temperature_comparison} in Appendix A.4 present an ablation study showing the effect of tuning $\alpha$ and $T$ independently. Each experiment was run with 5 random seeds.

\textbf{Results.} Table~\ref{tab:combined_hint_results} reports top-1 accuracy averaged across the three hint types (spatial, photometric, and cropping), with per-hint results provided in the Appendix. We also report the hint loss computed on virtual examples using the symmetric KL divergence with temperature $T=1$. The baseline hint loss is computed by evaluating the trained baseline model on virtual examples and their transformations, measuring the implicit invariance learned through augmentation alone.

\begin{table}[t]
  \caption{Top-1 accuracy (Acc.) and hint loss (Hint L.) on virtual examples for ViT-B and Swin-B models. Results averaged over spatial, photometric, 
  and cropping hints, with mean across 5 seeds. Baseline hint loss measures implicit invariance from augmentation; hints explicitly
  enforce invariance on virtual examples.}
  \label{tab:combined_hint_results}
  \begin{center}
    \scriptsize
    \setlength{\tabcolsep}{7pt}
    \begin{sc}
      \begin{tabular}{l cc cc cc cc}
        \toprule
        Dataset
          & \multicolumn{2}{c}{ViT-B} & \multicolumn{2}{c}{ViT-B w/ Hints} 
          & \multicolumn{2}{c}{Swin-B} & \multicolumn{2}{c}{Swin-B w/ Hints} \\
        \cmidrule(lr){2-3} \cmidrule(lr){4-5} \cmidrule(lr){6-7} \cmidrule(lr){8-9}
        & Acc. & Hint L. & Acc. & Hint L. & Acc. & Hint L. & Acc. & Hint L. \\
        \midrule
        Stanford Cars  & 90.15 & 0.714 & \textbf{90.77} & \textbf{4.4e-05} 
                       & 92.30 & 0.749 & \textbf{92.95} & \textbf{2.3e-07} \\
        FGVC Aircrafts & 83.77 & 0.722 & \textbf{85.09} & \textbf{1.8e-07} 
                       & 91.17 & 0.772 & \textbf{91.53} & \textbf{1.9e-07} \\
        CUB-200-2011  & 88.09 & 0.571 & \textbf{88.75} & \textbf{4.4e-05} 
                       & 90.30 & 0.460 & \textbf{90.88} & \textbf{2.9e-07} \\
        Oxford Flowers & 98.99 & 0.196 & \textbf{99.46} & \textbf{8.5e-04} 
                       & 99.60 & 0.176 & \textbf{99.67} & \textbf{3.8e-06} \\
        \bottomrule
      \end{tabular}
    \end{sc}
  \end{center}
  \vskip -0.25in
\end{table}

We observe consistent improvements across all datasets and both architectures through the use of generative hints. The substantial reduction in hint loss on virtual examples (from $\sim$0.2-0.7 down to $\sim$10$^{-7}$-10$^{-4}$) demonstrates that explicit invariance enforcement successfully teaches the model to respect the specified transformations. While data augmentation alone provides some implicit invariance (as reflected in the baseline hint loss), the explicit hint objective on virtual examples substantially improves both invariance alignment and classification performance. This validates our hypothesis that virtual examples provide additional learning signal beyond what is available through augmentation on labeled data alone.

\subsection{Medical Imaging: CheXpert}

To evaluate the robustness and generality of our approach, we extended our experiments to medical imaging using the CheXpert dataset \citep{irvin2019chexpert}. CheXpert is a large-scale chest radiograph dataset collected from Stanford Hospital, containing 224,316 X-rays from 65,240 patients, annotated for 14 common thoracic pathologies. For our experiments, we used 9 categories: No Finding, Enlarged Cardiomediastinum, Cardiomegaly, Lung Opacity, Pneumonia, Pleural Effusion, Pleural Other, Fracture, and Support Devices. Labels are automatically extracted from radiology reports using a rule-based NLP system.

\textbf{Generative Model} We trained StyleGAN3 at $256 \times 256$ on the full dataset (batch size 16, lr$_G$=0.0025, lr$_D$=0.001, $\gamma=8.0$), achieving FID 4.38 for 5000 kimgs (thousand images processed) and selected the checkpoint with the best FID.

\textbf{Classification Setup} We used ResNet50 with ImageNet pretraining, $256 \times 256$ grayscale inputs, MSE loss for both classification and hints, batch size 64, Adam optimizer (lr=$10^{-5}$), 5 epochs. Baseline used spatial augmentation (translation/rotation 0-5\%); hints added spatial invariance enforcement ($\alpha=0.1$) using the transformations on virtual examples. We used only spatial invariance for this dataset, as photometric hints are inappropriate for X-rays (brightness/contrast variations can be clinically meaningful) and crop hints risk removing diagnostic information. Each experiment was run for 5 random seeds.

\textbf{Results} Table~\ref{tab:chexpert_classification_results} reports classification MSE loss for each pathology. Generative hints improve performance on 7 out of 9 pathologies, with an average MSE reduction of 1.286\%. While two categories (No Finding, Pleural Effusion) show slight increases in MSE, the overall trend demonstrates the benefit of explicit invariance enforcement. The improvements are particularly notable for Support Devices (3.154\%), Pneumonia (2.008\%), and Enlarged Cardiomediastinum (2.086\%). These results demonstrate that generative hints generalize across domains (natural images to medical imaging), architectures (transformers to CNNs), and objectives (cross-entropy to MSE regression).

\begin{table}[t]
  \caption{Classification MSE loss across pathologies on CheXpert, with and without generative hints. Mean across 5 seeds. \textit{Percent Gain} represents the relative MSE reduction from baseline.}
  \label{tab:chexpert_classification_results}
  \begin{center}
    \scriptsize
    \setlength{\tabcolsep}{3pt} 
    \renewcommand{\arraystretch}{1.1} 
    \begin{sc}
      \begin{tabular}{l c c c}
        \toprule
        Pathology & Baseline & w/ Hints & \% Gain \\
        \midrule
        No Finding                & \textbf{0.636} & 0.639 & -0.472\% \\
        Enlarged Cardiomediastinum & 0.719 & \textbf{0.704} & 2.086\% \\
        Cardiomegaly              & 0.339 & \textbf{0.337} & 0.590\% \\
        Lung Opacity              & 0.795 & \textbf{0.784} & 1.384\% \\
        Pneumonia                 & 0.797 & \textbf{0.781} & 2.008\% \\
        Pleural Effusion          & \textbf{0.423} & 0.425 & -0.473\% \\
        Pleural Other             & 0.876 & \textbf{0.864} & 1.370\% \\
        Fracture                  & 0.673 & \textbf{0.660} & 1.932\% \\
        Support Devices           & 0.983 & \textbf{0.952} & 3.154\% \\
        \midrule
        \textbf{Average Gain}          & & & 1.286 \% \\
        \bottomrule
      \end{tabular}
    \end{sc}
  \end{center}
  \vskip -0.3in
\end{table}

\subsection{Effect of Generator Quality on Hint Learning}

\begin{figure}[h]
    \centering
    \includegraphics[width=0.7\linewidth]{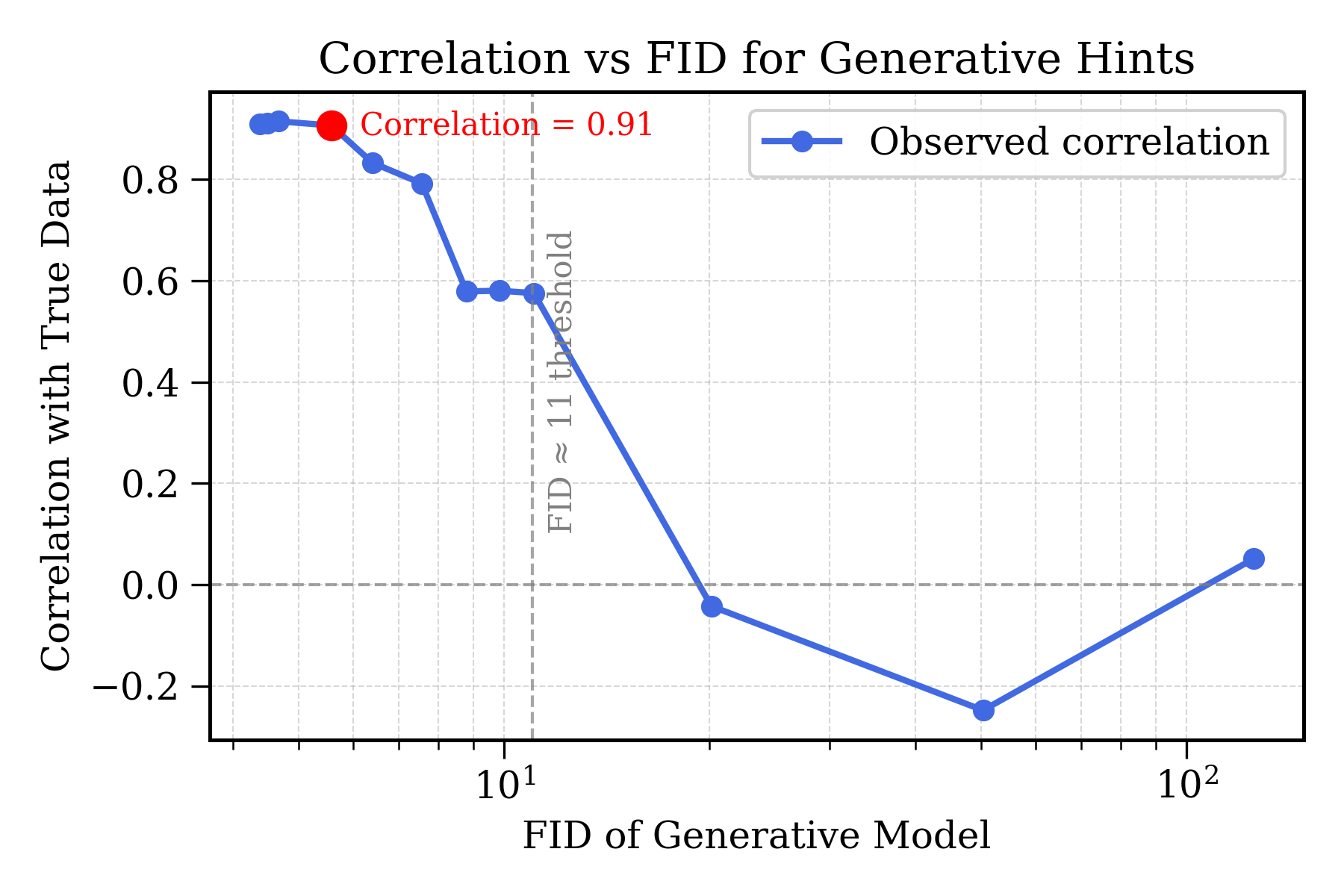}
    \vspace{-4pt}
    \caption{
        Correlation between generative hint loss on virtual examples and hint loss on real training data, plotted against FID. The horizontal dashed line marks zero correlation; the vertical dashed line marks the FID threshold ($\sim$11) where the generative model begins providing meaningful signal. The red point marks FID 5.58, where correlation reaches 0.91.
    }
    \vspace{-8pt}
    \label{fig:correlation_vs_FID}
\end{figure}

We conducted experiments to determine the generative model quality required for effective hint learning. Specifically, we identified the FID threshold where the generative model sufficiently captures the input distribution such that hints learned on virtual examples transfer to real training data.

\textbf{Experimental Setup.} Following the CheXpert setup from Section~5.3, we trained classification models using generative hints with generators of varying quality (FID scores). Crucially, models were trained \emph{only} on virtual examples using the hint loss (without any data augmentation on real training data), allowing us to isolate how well the generative model captures the true data distribution. We computed the correlation between hint loss on virtual examples and hint loss on real training data across 120 checkpoints sampled throughout 5 epochs of training. High correlation indicates that learning on virtual examples reflects learning on real data.

\textbf{Results.} Figure~\ref{fig:correlation_vs_FID} shows correlation versus FID. At FID values above 50, correlation is near zero, indicating that low-quality generative models provide no meaningful learning signal. Once FID drops below 11, correlation becomes substantial, reaching 0.91 at FID 5.58. Beyond this point, further improvements in FID yield diminishing returns in correlation.

To understand how generator quality affects final classification performance, we ran an additional ablation using Swin-B with the photometric hint on Stanford Cars (Table~\ref{tab:fid_accuracy_comp}). Performance remains stable and beneficial when FID is below $\sim$12, but degrades once FID exceeds 15. These results indicate that while high-quality generative models ($\text{FID} < 10$) are ideal, moderate-quality models ($10 < \text{FID} < 15$) still provide meaningful benefits for hint learning. The full table can be observed in Appendix A.5.

\section{Conclusion}
We proposed a method to reformulate supervised classification as semi-supervised learning by treating data synthesized from a generative model as unlabeled data, enabling models to learn functional properties through virtual examples. Our evaluations across fine-grained visual classification and medical imaging domains showed that generative hints consistently outperformed traditional data augmentation when learning the same properties, without requiring perfect generative models. Future work includes developing dynamic schedulers to adapt objective weights, designing embedding hints over latent representations for properties difficult to encode via augmentation, and extending to object detection and segmentation tasks. This work establishes generative hints as a versatile tool for domain knowledge injection, opening new avenues for explicit regularization in fully labeled settings.

\newpage
\bibliography{iclr2026_delta}

@article{AbuMostafa1995Hints,
  author    = {Yaser S. Abu-Mostafa},
  title     = {Hints},
  journal   = {Neural Computation},
  volume    = {7},
  pages     = {639--671},
  month     = jul,
  year      = {1995}
}

@inproceedings{goodfellow2014generative,
  title={Generative Adversarial Nets},
  author={Goodfellow, Ian J. and Pouget-Abadie, Jean and Mirza, Mehdi and Xu, Bing and Warde-Farley, David and Ozair, Sherjil and Courville, Aaron and Bengio, Yoshua},
  booktitle={Advances in Neural Information Processing Systems},
  volume={27},
  year={2014},
  publisher={Curran Associates, Inc.}
}

@inproceedings{ho2020denoising,
  title={Denoising Diffusion Probabilistic Models},
  author={Ho, Jonathan and Jain, Ajay and Abbeel, Pieter},
  booktitle={Advances in Neural Information Processing Systems},
  volume={33},
  pages={6840--6851},
  year={2020},
  publisher={Curran Associates, Inc.}
}

@inproceedings{rombach2022high,
  title={High-Resolution Image Synthesis with Latent Diffusion Models},
  author={Rombach, Robin and Blattmann, Andreas and Lorenz, Dominik and Esser, Patrick and Ommer, Bjorn},
  booktitle={Proceedings of the IEEE/CVF Conference on Computer Vision and Pattern Recognition (CVPR)},
  year={2022},
  pages={10684--10695}
}

@inproceedings{song2020score,
  title={Score-Based Generative Modeling through Stochastic Differential Equations},
  author={Song, Yang and Meng, Jia and Ermon, Stefano},
  booktitle={International Conference on Learning Representations (ICLR)},
  year={2021},
  url={https://arxiv.org/abs/2011.13456}
}

@inproceedings{karras2019stylegan,
  title={A Style-Based Generator Architecture for Generative Adversarial Networks},
  author={Karras, Tero and Laine, Samuli and Aila, Timo},
  booktitle={Proceedings of the IEEE/CVF Conference on Computer Vision and Pattern Recognition (CVPR)},
  pages={4401--4410},
  year={2019}
}

@inproceedings{karras2020stylegan,
  title={Analyzing and Improving the Image Quality of StyleGAN},
  author={Karras, Tero and Laine, Samuli and Aittala, Miika and Hellsten, Janne and Lehtinen, Jaakko and Aila, Timo},
  booktitle={Proceedings of the IEEE/CVF Conference on Computer Vision and Pattern Recognition (CVPR)},
  pages={8110--8119},
  year={2020}
}

@inproceedings{karras2021stylegan,
  title={Alias-Free Generative Adversarial Networks},
  author={Karras, Tero and Laine, Samuli and Aittala, Miika and Hellsten, Janne and Lehtinen, Jaakko and Aila, Timo},
  booktitle={Advances in Neural Information Processing Systems (NeurIPS)},
  volume={34},
  pages={852--863},
  year={2021}
}

@inproceedings{Karras2020ada,
  title     = {Training Generative Adversarial Networks with Limited Data},
  author    = {Tero Karras and Miika Aittala and Janne Hellsten and Samuli Laine and Jaakko Lehtinen and Timo Aila},
  booktitle = {Proc. NeurIPS},
  year      = {2020}
}

@inproceedings{radford2015unsupervised,
  title={Unsupervised Representation Learning with Deep Convolutional Generative Adversarial Networks},
  author={Radford, Alec and Metz, Luke and Chintala, Soumith},
  booktitle={International Conference on Learning Representations (ICLR)},
  year={2016},
  url={https://arxiv.org/abs/1511.06434}
}

@inproceedings{vaswani2017attention,
  title={Attention is All You Need},
  author={Vaswani, Ashish and Shazeer, Noam and Parmar, Niki and Uszkoreit, Jakob and Jones, Llion and Gomez, Aidan N and Kaiser, Łukasz and Polosukhin, Illia},
  booktitle={Advances in Neural Information Processing Systems (NeurIPS)},
  pages={5998--6008},
  year={2017},
  publisher={Curran Associates, Inc.},
  url={https://arxiv.org/abs/1706.03762}
}

@inproceedings{resnet,
  title={Deep Residual Learning for Image Recognition},
  author={He, Kaiming and Zhang, Xiangyu and Ren, Shaoqing and Sun, Jian},
  booktitle={Proceedings of the IEEE Conference on Computer Vision and Pattern Recognition (CVPR)},
  pages={770--778},
  year={2016}
}

@inproceedings{vit,
  title={An Image is Worth 16x16 Words: Transformers for Image Recognition at Scale},
  author={Dosovitskiy, Alexey and Beyer, Lucas and Kolesnikov, Alexander and Weissenborn, Dirk and Zhai, Xiaohua and Unterthiner, Thomas and Dehghani, Mostafa and Minderer, Matthias and Heigold, Georg and Gelly, Sylvain and Uszkoreit, Jakob and Houlsby, Neil},
  booktitle={International Conference on Learning Representations (ICLR)},
  year={2021}
}

@inproceedings{swint,
  title={Swin Transformer: Hierarchical Vision Transformer using Shifted Windows},
  author={Liu, Ze and Lin, Yutong and Cao, Yue and Hu, Han and Wei, Zhuliang and Zhang, Zheng and Lin, Stephen and Guo, Baining},
  booktitle={Proceedings of the IEEE/CVF International Conference on Computer Vision (ICCV)},
  pages={10012--10022},
  year={2021}
}

@article{AbuMostafa1990Hints,
  title        = {Learning from Hints in Neural Networks},
  author       = {Abu-Mostafa, Yaser S.},
  journal      = {Journal of Complexity},
  volume       = {6},
  number       = {2},
  pages        = {192--198},
  year         = {1990},
  doi          = {10.1016/0885-064X(90)90024-4},
  publisher    = {Elsevier}
}

@inproceedings{heusel2017gans,
  title={GANs Trained by a Two Time-Scale Update Rule Converge to a Local Nash Equilibrium},
  author={Heusel, Martin and Ramsauer, Hubert and Unterthiner, Thomas and Nessler, Bernhard and Hochreiter, Sepp},
  booktitle={Advances in Neural Information Processing Systems},
  volume={30},
  pages={6626--6637},
  year={2017}
}

@inproceedings{kingma2014semi,
  title={Semi-supervised learning with deep generative models},
  author={Kingma, Diederik P and Rezende, Danilo J and Mohamed, Shakir and Welling, Max},
  booktitle={Advances in Neural Information Processing Systems (NeurIPS)},
  pages={3581--3589},
  year={2014},
  url={https://papers.nips.cc/paper/5352-semi-supervised-learning-with-deep-generative-models}
}

@inproceedings{imagenet,
  title={ImageNet: A Large-Scale Hierarchical Image Database},
  author={Deng, Jia and Dong, Wei and Socher, Richard and Li, Li-Jia and Li, Kai and Fei-Fei, Li},
  booktitle={2009 IEEE Conference on Computer Vision and Pattern Recognition (CVPR)},
  pages={248--255},
  year={2009},
  organization={IEEE}
}

@article{azizi2023synthetic,
  title={Synthetic data from diffusion models improves ImageNet classification},
  author={Azizi, Shekoofeh and Kornblith, Simon and Saharia, Chitwan and Norouzi, Mohammad and Fleet, David J},
  journal={Transactions on Machine Learning Research},
  volume={4},
  pages={1--15},
  year={2023},
  url={https://arxiv.org/abs/2304.08466}
}

@inproceedings{yuan2024realfake,
  title={Real-Fake: Effective Training Data Synthesis Through Distribution Matching},
  author={Jianhao Yuan and Jie Zhang and Shuyang Sun and Philip Torr and Bo Zhao},
  booktitle={Proceedings of the International Conference on Learning Representations (ICLR)},
  year={2024},
  url={https://openreview.net/forum?id=svIdLLZpsA}
}

@inproceedings{irvin2019chexpert,
  title={CheXpert: A Large Chest Radiograph Dataset with Uncertainty Labels and Expert Comparison},
  author={Jeremy Irvin and Pranav Rajpurkar and Michael Ko and Yifan Yu and Silviana Ciurea-Ilcus and Chris Chute and Henrik Marklund and Behzad Haghgoo and Robyn Ball and Katie Shpanskaya and Jayne Seekins and David A. Mong and Safwan S. Halabi and Jesse K. Sandberg and Ricky Jones and David B. Larson and Curtis P. Langlotz and Bhavik N. Patel and Matthew P. Lungren and Andrew Y. Ng},
  booktitle={Proceedings of the AAAI Conference on Artificial Intelligence},
  volume={33},
  pages={590--597},
  year={2019},
  doi={10.1609/aaai.v33i01.3301590},
  url={https://doi.org/10.1609/aaai.v33i01.3301590},
  note={\url{https://stanfordmlgroup.github.io/competitions/chexpert/}}
}

@inproceedings{krause2013collecting,
  title={Collecting a Large-Scale Dataset of Fine-Grained Cars},
  author={Jonathan Krause and Jia Deng and Michael Stark and Li Fei-Fei},
  booktitle={Proceedings of the Second Workshop on Fine-Grained Visual Categorization (FGVC2)},
  year={2013},
  url={https://ai.stanford.edu/~jkrause/papers/fgvc13.pdf},
  note={\url{https://ai.stanford.edu/~jkrause/papers/fgvc13.pdf}}
}

@techreport{WahCUB_200_2011,
  title = {The Caltech-UCSD Birds-200-2011 Dataset},
  author = {Wah, C. and Branson, S. and Welinder, P. and Perona, P. and Belongie, S.},
  year = {2011},
  institution = {California Institute of Technology},
  number = {CNS-TR-2011-001},
  url = {http://www.vision.caltech.edu/visipedia/CUB-200-2011.html}
}

@techreport{maji13fine-grained,
  title         = {Fine-Grained Visual Classification of Aircraft},
  author        = {Subhransu Maji and Juho Kannala and Esa Rahtu and Matthew Blaschko and Andrea Vedaldi},
  year          = {2013},
  institution   = {arXiv},
  number        = {1306.5151},
  archivePrefix = {arXiv},
  eprint        = {1306.5151},
  primaryClass  = {cs.CV},
  url           = {https://arxiv.org/abs/1306.5151}
}

@inproceedings{nilsback2008flowers,
  title={Automated flower classification over a large number of classes},
  author={Nilsback, Maria-Elena and Zisserman, Andrew},
  booktitle={Proceedings of the 6th Indian Conference on Computer Vision, Graphics and Image Processing (ICVGIP)},
  pages={722--729},
  year={2008},
  organization={ACM},
  url={https://www.robots.ox.ac.uk/~vgg/data/flowers/102/}
}

@inproceedings{antoniou2017dagan,
  title={Data Augmentation Generative Adversarial Networks},
  author={Antoniou, Antreas and Storkey, Amos and Edwards, Harrison},
  booktitle={arXiv preprint arXiv:1711.04340},
  year={2017}
}

@inproceedings{fridadar2018liver,
  title={GAN-based synthetic medical image augmentation for increased CNN performance in liver lesion classification},
  author={Frid-Adar, Maayan and Klang, Eyal and Amitai, Michal and Goldberger, Jacob and Greenspan, Hayit},
  booktitle={arXiv preprint arXiv:1803.01229},
  year={2018}
}

@inproceedings{zhao2019transform,
  title={Data augmentation using learned transformations for one-shot medical image segmentation},
  author={Zhao, Hang and Li, Bowen and Xu, Tao and et al.},
  booktitle={arXiv preprint arXiv:1902.09383},
  year={2019}
}

@inproceedings{bordes2023diffusion,
  title={Synthetic Data from Diffusion Models Improves ImageNet Classification},
  author={Bordes, Florian and Mu, Norman and Faltings, Boi and et al.},
  booktitle={arXiv preprint arXiv:2304.08466},
  year={2023}
}

@inproceedings{huang2023ttida,
  title={TTIDA: Controllable Generative Data Augmentation via Text-to-Text and Text-to-Image Models},
  author={Huang, Zixuan and Yang, Yuchen and Liang, Chen and et al.},
  booktitle={arXiv preprint arXiv:2304.08821},
  year={2023}
}

@inproceedings{zhang2024aga,
  title={Data Augmentation for Image Classification using Generative AI},
  author={Zhang, Kai and Wang, Ming and Liu, Jian and et al.},
  booktitle={arXiv preprint arXiv:2409.00547},
  year={2024}
}

@inproceedings{mahapatra2022unified,
  title={Unified Framework for Histopathology Image Augmentation and Classification via Generative Models},
  author={Mahapatra, Dwarikanath and Ge, Zongyuan},
  booktitle={arXiv preprint arXiv:2212.09977},
  year={2022}
}

@article{perez2017effectiveness,
  title={The effectiveness of data augmentation in image classification using deep learning},
  author={Perez, Luis and Wang, Jason},
  journal={CoRR},
  volume={abs/1712.04621},
  year={2017},
  url={http://arxiv.org/abs/1712.04621},
  eprinttype={arxiv},
  eprint={1712.04621}
}

@article{shorten2019survey,
  title={A survey on image data augmentation for deep learning},
  author={Shorten, Connor and Khoshgoftaar, Taghi M.},
  journal={Journal of Big Data},
  volume={6},
  number={1},
  pages={60},
  year={2019},
  publisher={Springer},
  doi={10.1186/s40537-019-0197-0}
}
\bibliographystyle{iclr2026_delta}

\newpage
\appendix
\section{Appendix}
\subsection{Datasets}

We ran on the 4 datasets Stanford Cars, FGVC Aircrafts, CUB-200-2011, and Oxford Flowers. The datasets are all fine grain visual classification datasets with 100 or more classes. Full dataset specifications can be seen in Table 2. For datasets with a training/val/test split we combined the training and validation set, which is considered the standard for the datasets. 

\begin{table}[h]
    \caption{Summary statistics for fine-grained visual classification datasets: number of classes, total image count, and standard train/test splits.}
    \captionsetup{skip=10pt}
    \centering
    \renewcommand{\arraystretch}{1.2}
    \begin{tabular}{l c c c c}
        \toprule
        Dataset             & \# Classes & Total Images & Train Images & Test Images \\
        \midrule
        Stanford Cars       & 196        & 16,185       & 8,144        & 8,041        \\
        FGVC Aircrafts      & 100        & 10,200       & 6,800 & 3,400 \\
        CUB-200-2011        & 200        & 11,788       & 5,994 & 5,794 \\
        Oxford Flowers 102  & 102        & 8,189 & 2,040        & 6,149        \\
        \bottomrule
    \end{tabular}
    \label{tab:dataset-stats}
\end{table}

\subsection{Generative Model Training}

Generative models were trained according to the specifications listed in Table~\ref{tab:generative-hyperparams}, using only the training split of each dataset. 
All models were trained until convergence with Adaptive Discriminator Augmentation (ADA)~\cite{Karras2020ada}, in which augmentations are applied to images before being passed to the discriminator. 
Training was performed on a single NVIDIA H100 GPU and continued until Fréchet Inception Distance (FID) convergence~\cite{heusel2017gans}. The augmentations used included: \texttt{xflip}, \texttt{rotate90}, \texttt{xint}, \texttt{scale}, \texttt{rotate}, \texttt{anisco}, \texttt{xfrac}, \texttt{brightness}, \texttt{contrast}, \texttt{lumaflip}, \texttt{hue}, and \texttt{saturation}. 
Full augmentation specifications are available in the official StyleGAN3 repository. The resulting generative model FIDs can be observed in Table~\ref{tab:dataset-stats}.

\begin{table}[h]
\centering
\captionsetup{skip=10pt}
\caption{Training hyperparameters for the generative model (StyleGAN3 with ADA).}
\begin{tabular}{l l}
\toprule
\textbf{Hyperparameter} & \textbf{Value} \\
\midrule
Model Type              & StyleGAN3 \\
Resolution        & 512 × 512 \\
Adaptive Discriminator Augmentation & Enabled \\
Mirror                  & Enabled \\
Optimizer               & AdamW \\
Generator Learning Rate           & 0.0025 \\
Discriminator Learning Rate           & 0.001 \\
Batch Size              & 16 \\
Gamma                   & 4.0 \\
Stop Condition        & FID convergence \\

\bottomrule
\end{tabular}
\label{tab:generative-hyperparams}
\end{table}

\begin{table}[h]
    \captionsetup{skip=10pt}
    \caption{The resulting StyleGAN3 models trained on each of the datasets including the FID achieved.}
    \centering
    \renewcommand{\arraystretch}{1.2}
    \begin{tabular}{l c c c c}
        \toprule
        Dataset             & \# Classes & Train Images & FID \\
        \midrule
        Stanford Cars       & 196 & 8,144  & 4.27 \\
        FGVC Aircrafts      & 100        & 6,800 & 4.72 \\
        CUB-200-2011        & 200 & 5,994 & 7.37 \\
        Oxford Flowers 102  & 102 & 2,040 & 12.62 \\
        \bottomrule
    \end{tabular}
    \label{tab:dataset-stats-appendix}
\end{table}

\subsection{Training Hyperparameters and Implementation Details}

Both the ViT-B/16 and Swin-B transformer models were pretrained on ImageNet~\cite{imagenet}. Table~\ref{tab:model-hyperparams} reports the training hyperparameters, which were tuned to maximize baseline performance (without hints) and then kept fixed when applying generative hints. The only parameters modified for hint-based training were the hint loss weight $\alpha$ and the temperature $T$ used in the symmetric KL divergence loss.

\textbf{Data Augmentation.} For baseline training (without hints), we applied the following augmentations: horizontal flip with probability $p=0.5$, random rotation uniformly sampled from $[0^{\circ}, 5^{\circ}]$, and random translation uniformly sampled from $[0\%, 5\%]$ of image dimensions. When applying generative hints, we used identical augmentations on the training data, except that horizontal flip probability was increased to $p=1.0$ for the spatial invariance hint to enforce complete flip invariance.

\textbf{Hint-Specific Transformations.} The three hint types used the following transformations:
\begin{itemize}
    \item \textbf{Spatial Invariance:} Horizontal flip ($p=1.0$), rotation $[0^{\circ}, 5^{\circ}]$, translation $[0\%, 5\%]$
    \item \textbf{Photometric Invariance:} Brightness, contrast, and saturation adjusted by factors uniformly sampled from $[0.8, 1.2]$ (i.e., $\pm 20\%$)
    \item \textbf{Cropping Invariance:} Images resized to $448 \times 448$ and randomly cropped to $384 \times 384$
\end{itemize}

\textbf{Virtual Example Generation.} During hint-based training, we generated one virtual example for each real sample per training batch by sampling from the pretrained StyleGAN3 model. Images were generated at $512 \times 512$ resolution and resized to $384 \times 384$ to match the model input size. We alternated between optimizing the supervised cross-entropy loss on real training data and the hint loss on virtual examples at every batch.

\textbf{Hint Loss Configuration.} We used symmetric KL divergence with temperature $T=0.8$ for all experiments. Table~\ref{tab:model-hyperparams} reports the best-performing $\alpha$ for each architecture, which was then used consistently across all datasets and hint types for that architecture. All training was performed on a single NVIDIA H100 GPU.

\begin{table}[h]
\centering
\captionsetup{skip=10pt}
\caption{Training and model hyperparameters for ViT-B/16 and Swin-B.}
\begin{tabular}{l c c}
\toprule
\textbf{Hyperparameter} & \textbf{ViT-B/16} & \textbf{Swin-B} \\
\midrule
Resolution              & $384\times 384$ & $384 \times 384$ \\
Optimizer               & AdamW & AdamW \\
Learning Rate           & $1\mathrm{e}{-4}$ & $1\mathrm{e}{-4}$ \\
Weight Decay            & 0.01 & 0.01 \\
Batch Size              & 32 & 32 \\
Scheduler               & Cosine Annealing & Cosine Annealing \\
Number of Epochs        & 200 & 200 \\
Hint Loss Weight        & 1.0 & 50.0 \\
\bottomrule
\end{tabular}
\label{tab:model-hyperparams}
\end{table}

\begin{table}[h]
    \caption{
            Top-1 accuracy using the photometric invariance hint for Stanford Cars, FGVC Aircraft, CUB-200-2011, and Oxford Flowers. Mean across 5 seeds. Bold indicates best performance. Average improvement: 0.76\% (max: 2.10\%).
        }
    \centering
    \setlength{\tabcolsep}{8pt} 
    \begin{tabular}{l c c c c}
        \toprule
        Dataset 
            & ViT-B Baseline
            & ViT-B w/ Hints 
            & Swin-B Baseline
            & Swin-B w/ Hints \\
        \midrule
        Stanford Cars   & 89.15 & \textbf{90.29} & 91.11 & \textbf{92.13} \\
        FGVC Aircrafts  & 82.51 & \textbf{84.61} & 90.23 & \textbf{90.66} \\
        CUB-200-2011   & 87.92 & \textbf{88.37} & 90.06 & \textbf{90.58} \\
        Oxford Flowers  & 99.14 & \textbf{99.45} & 99.58 & \textbf{99.66} \\
        \bottomrule
    \end{tabular}
    
    \label{tab:hint_results_transposed}
\end{table}

\begin{table}[h]
    \caption{
            Top-1 accuracy using the cropping invariance hint for Stanford Cars, FGVC Aircraft, CUB-200-2011, and Oxford Flowers. Mean across 5 seeds. Bold indicates best performance. Average improvement: 0.46\% (max: 1.20\%).
            Top-1 accuracy using the Crop hint for the Stanford Cars, FGVC Aircraft, CUB-200-2011, and Oxford Flowers datasets. Bold indicates the best performance for each dataset and model. Hints results in an average improvement of 0.38\% (up to 1.2\%).
        }
    \centering
    \setlength{\tabcolsep}{8pt} 
    \begin{tabular}{l c c c c}
        \toprule
        Dataset 
            & ViT-B Baseline
            & ViT-B w/ Hints 
            & Swin-B Baseline
            & Swin-B w/ Hints \\
        \midrule
        Stanford Cars   & 90.39 & \textbf{90.43} & 92.87 & \textbf{93.18} \\
        FGVC Aircrafts  & 82.37 & \textbf{82.44} & 90.74 & \textbf{91.11} \\
        CUB-200-2011   & 87.91 & \textbf{89.11} & 90.57 & \textbf{90.96} \\
        Oxford Flowers  & 98.89 & \textbf{99.50} & 99.62 & \textbf{99.67} \\
        \bottomrule
    \end{tabular}
    
    \label{tab:hint_results_transposed}
\end{table}

\begin{table}[h]
    \caption{
            Top-1 accuracy using the spatial invariance hint for Stanford Cars, FGVC Aircraft, CUB-200-2011, and Oxford Flowers. Mean across 5 seeds. Bold indicates best performance. Average improvement: 0.63\% (max: 1.78\%).
        }
    \centering
    \setlength{\tabcolsep}{8pt} 
    \begin{tabular}{l c c c c}
        \toprule
        Dataset 
            & ViT-B Baseline
            & ViT-B w/ Hints 
            & Swin-B Baseline
            & Swin-B w/ Hints \\
        \midrule
        Stanford Cars   & 90.90 & \textbf{91.58} & 92.92 & \textbf{93.53} \\
        FGVC Aircrafts  & 86.43 & \textbf{88.21} & 92.55 & \textbf{92.83} \\
        CUB-200-2011   & 88.45 & \textbf{88.76} & 90.28 & \textbf{91.11} \\
        Oxford Flowers  & 98.94 & \textbf{99.43} & 99.61 & \textbf{99.68} \\
        \bottomrule
    \end{tabular}
    \label{tab:hint_results_transposed}
\end{table}

\textbf{Across all datasets and architectures, we observe:}

\begin{enumerate}
    \item \textbf{Photometric hints provide the largest gains} (avg. 0.76\%), particularly on FGVC Aircraft where color/lighting variations are less semantically meaningful than spatial structure.
    
    \item \textbf{Spatial hints provide consistent improvements} (avg. 0.63\%) across all datasets, demonstrating that explicit enforcement of flip and rotation invariance provides signal beyond standard augmentation.
    
    \item \textbf{Cropping hints show more modest gains} (avg. 0.38\%), likely because cropping can remove discriminative details in fine-grained classification, making strict invariance less appropriate.
    
    \item \textbf{All three hints improve over baseline}, indicating that virtual examples successfully provide additional invariance signal regardless of the specific transformation type.
    
    \item \textbf{Improvements are consistent across architectures}, with both ViT-B and Swin-B benefiting from hints, demonstrating generalizability.
\end{enumerate}

The variability in per-hint performance suggests that hint selection could be optimized per-dataset based on domain knowledge about which invariances are most appropriate.

\subsection{Alpha and Temperature Sensitivity Analysis}

To understand the robustness of generative hints to hyperparameter choices, we conducted ablation studies on the two key parameters of our method: the hint loss weight $\alpha$ and the temperature parameter $T$ used in the symmetric KL divergence loss. Both experiments were conducted using Swin-B on the Stanford Cars dataset with the photometric invariance hint, keeping all other hyperparameters fixed as specified in Table~\ref{tab:model-hyperparams}.

\subsubsection{Effect of Hint Loss Weight $\alpha$}

The hint loss weight $\alpha$ controls the relative importance of the invariance objective compared to the supervised classification objective. Table~\ref{tab:alpha_comparison} shows classification accuracy as a function of $\alpha$. We observe that hints provide consistent improvements over the baseline (91.11\%) across a wide range of $\alpha$ values from 1.0 to 100.0, with peak performance at $\alpha = 50.0$ (92.13\%). 

\begin{table}[h]
    \caption{
            Classification accuracy as a function of hint loss weight $\alpha$ for Swin-B on Stanford Cars with photometric hint. Baseline corresponds to training without hints. The method is robust across two orders of magnitude.
        }
    \centering
    \setlength{\tabcolsep}{10pt}
    \begin{tabular}{l c c}
        \toprule
        Setting & $\alpha$ & Accuracy \\
        \midrule
        Baseline (No Hints) & -- & 91.11 \\
        \midrule
        w/Hints & 1.0   & 91.52 \\
        w/Hints & 5.0   & 91.77 \\
        w/Hints & 10.0  & 91.85 \\
        w/Hints & 25.0  & 91.63 \\
        w/Hints & 50.0  & 92.13 \\
        w/Hints & 75.0  & 91.84 \\
        w/Hints & 100.0 & 92.02 \\
        \bottomrule
    \end{tabular}
    \label{tab:alpha_comparison}
\end{table}

\subsubsection{Effect of Temperature Parameter $T$}

The temperature parameter $T$ controls the sharpness of the probability distributions in the symmetric KL divergence loss. Lower temperatures produce sharper distributions, while higher temperatures produce smoother distributions. Table~\ref{tab:temperature_comparison} shows classification accuracy as a function of $T$.

\begin{table}[h]
    \caption{
            Classification accuracy as a function of temperature $T$ for Swin-B on Stanford Cars with photometric hint. Baseline corresponds to training without hints. Moderate temperatures ($T \in [0.5, 1.2]$) provide stable performance.
        }
    \centering
    \setlength{\tabcolsep}{10pt}
    \begin{tabular}{l c c}
        \toprule
        Setting & $T$ & Accuracy \\
        \midrule
        Baseline (No Hints) & -- & 91.11 \\
        \midrule
        w/Hints & 0.5 & 92.05 \\
        w/Hints & 0.8 & 92.13 \\
        w/Hints & 1.0 & 91.80 \\
        w/Hints & 1.2 & 91.90 \\
        w/Hints & 1.5 & 91.58 \\
        w/Hints & 2.0 & 91.51 \\
        \bottomrule
    \end{tabular}
    \label{tab:temperature_comparison}
\end{table}

\subsection{Generator Quality Ablation}

To investigate how the quality of the generator affects downstream classification, we vary the generator’s FID and measure its impact on accuracy using photometric hints with Swin-B on Stanford Cars. Table~\ref{tab:fid_accuracy_comp} shows that hints remain beneficial at moderate FID levels, with performance improving as generator quality increases, before plateauing or declining at lower quality levels.

\begin{table}[h]
    \caption{Effect of generator quality (FID) on classification accuracy using photometric hint with Swin-B on Stanford Cars. Mean across 5 seeds. Performance remains beneficial at moderate FID levels before declining as quality degrades.}
    \centering
    \setlength{\tabcolsep}{10pt}
    \begin{tabular}{l c c}
        \toprule
        Setting & FID $\downarrow$ & Accuracy (\%) $\uparrow$ \\
        \midrule
        Data Aug. (No Hints) & -- & 91.11 \\  
        \midrule
        w/ Hints & 30.83 & 91.09 \\
        w/ Hints & 19.15 & 91.33 \\
        w/ Hints & 14.49 & 91.59 \\
        w/ Hints & 11.68 & 91.83 \\
        w/ Hints & 5.29 & 92.13 \\    
        \bottomrule
    \end{tabular}
    \label{tab:fid_accuracy_comp}
\end{table}

\end{document}